\documentclass[runningheads]{llncs}
\usepackage{graphicx}
\usepackage{booktabs}
\usepackage{multirow}
\usepackage{soul,color}

\begin{document}
\title{Clutter Slices Approach for Identification-on-the-fly of Indoor Spaces}
%
%
%
\author{Upinder Kaur\inst{1}\thanks{equal contribution} \and
Praveen Abbaraju\inst{1}\inst{+}\inst{*} \and
Harrison McCarty\inst{1} \and
Richard M. Voyles\inst{1}}
\authorrunning{U. Kaur, P. Abbaraju et al.}
%
\institute{$^1$ Purdue University, West Lafayette, Indiana, 47907, USA \\
\email{\{kauru, pabbaraj, mccarth, rvoyles\}@purdue.edu} \\
\email{$^+$ Corresponding author} \\
}

\maketitle 

\begin{abstract}

Construction spaces are constantly evolving, dynamic environments in need of continuous surveying, inspection, and assessment. Traditional manual inspection of such spaces proves to be an arduous and time-consuming activity. Automation using robotic agents can be an effective solution. Robots, with perception capabilities can autonomously classify and survey indoor construction spaces. In this paper, we present a novel identification-on-the-fly approach for coarse classification of indoor spaces using the unique signature of clutter. Using the context granted by clutter, we recognize common indoor spaces such as corridors, staircases, shared spaces, and restrooms. The proposed clutter slices pipeline achieves a maximum accuracy of 93.6\% on the presented clutter slices dataset. This sensor independent approach can be generalized to various domains to equip intelligent autonomous agents in better perceiving their environment. 

\end{abstract}
\section{Introduction}

Large-scale construction spaces need periodic surveying, inspection, and renovation \cite{jupp20174d}. Continuous assessment helps to identify the completion status as well as localize problems which may arise \cite{becerik2012application}. Traditionally, this requires a well-coordinated surveillance activity which consumes enormous man-hours, even resulting in delays and overheads. Further, the inherent complexity of such spaces, in terms of design, inter-connectivity, and scale complicate this already arduous undertaking. Automation of processes in such activities has the potential to greatly reduce the effort required and boost overall productivity, at the same time reducing overhead costs and delays. This need for process automation in the complex and fast-paced world of construction calls for innovation at all levels. 

Inspection and surveying of outdoor large-scale construction activities now utilizes satellite imagery and Global Positioning Systems (GPS) based localization \cite{zhang2016indoor,aydin2011ins}. While these methods are robust and cost effective solutions for outdoor spaces, they prove to be in-effective for indoor spaces. Moreover, indoor GPS-based navigation is not effective for multi-level structures and the signal in itself becomes unreliable \cite{aydin2011ins}. Alternative solutions including WiFi signals and quantum sensors require expensive equipment for implementation \cite{mckinlay2016technology}. Further, these limitations of expenses, time, and resources for efficient inspection and surveillance is withholding the extensive application of Building Information Modelling (BIM) in construction activities \cite{becerik2012application,sun2017literature}. Robotic technologies, such as mobile robots, rovers, and aerial manipulators, are proving to be an efficient automation solution for construction activities \cite{constructionrobots}. Mobile robots, such as aerial manipulators (UAV) \cite{praveenICRA2019} and ground-based manipulators (wheeled and legged) \cite{wang2012smartguard} are a cost-effective solution for real-time large scale inspections due to their robust and reliable performance.   

Mobile robots with capabilities of perception are proving to be a paradigm shifting technology in inspection and surveillance. Perception sensors such as LiDARs (2D and 3D), stereo cameras, RGB-D cameras, ultrasonic and infrared proximity sensors have been extensively used in robot vision to identify the surrounding of a robot and its subsequent localization \cite{lidarslam2017,lidarslam2019}. This is similar to human perceiving their surroundings through multiple modal sensing. For example, humans use vision and their knowledge base to comprehend the characteristics of a construction site. They also use tactile sensing to provide an understanding over various states and properties of surfaces \cite{wang2012smartguard,wang2019}. However, humans have the ability to inherently perform these identification procedures as a secondary task, while performing primary targeted tasks such as reaching a target location, navigating among obstacles, etc. We call this identification-on-the-fly as it enables multi modal perception for intelligent and self-adaptive systems. \cite{praveenIROS2020}. Extending this methodology to coarse identification and classification of indoor spaces yields systems capable of multi-modal perception and intelligent operation yet efficient, especially for BIM development.  

In this paper, the identification-on-the-fly method is used to coarsely identify human-built spaces based on the distribution of clutter.  Each space has its own distinct signature. Clutter, the things which occupy space in an indoor environments such as doors, desks, and wall fittings, grant context to a space. The aim of this study is to develop the ability to identify and classify spaces based on this inherent signature. Hence, we present a unique sensor independent approach for classifying indoor spaces based on their inherent signatures. A sensor independent approach allows generalization of this method to numerous avenues and also allows for fast and inexpensive implementations. 

In order to develop and validate this approach, we first present the Clutter Slices dataset. This initial dataset is developed with 2D LiDAR scans of indoor areas, such as staircases, washrooms, corridors, and shared spaces; spaces are common to most developments. We then propose the clutter slices pipeline which utilizes commonly used classifiers to train and subsequently test the approach on the collected dataset. Hence, the contributions of this study are as follows:

\begin{itemize}
\item The Clutter Slices dataset of common indoor spaces along with the analysis of its distribution. This dataset is publicly available.

\item The clutter slices classification pipeline, including widely used classifiers, is presented. The evaluation of this model on the clutter slices dataset is presented as a baseline.

\item A new pipeline for clutter slices classification independent of sensor type, including widely used classifiers. The evaluation of this model on the clutter slices dataset is presented as a baseline.

\item Performance analysis of the selected classifiers in the proposed pipeline is presented on the clutter slices dataset. 
\end{itemize}

The organization of this paper is as follows: Section 2 describes the \\
Identification-on-the-fly approach using clutter slices to decipher the unique signatures of indoor spaces. Further, Section 3 presents the Clutter Slices dataset. In this section, we describe the methodology of data collection and the structure of the dataset. Section 4 presents the model and the classification methods used on the Clutter Slices dataset for identification of spaces. Experiments and results are presented in Section 5, followed by the conclusion in Section 6.   

\section{Identification-on-the-fly}

Embedding intelligence and self-adaptive features into robots requires them to perform multi-modal tasks, simultaneously, to extract a rich understanding of their environment. Such rich comprehension is based on contextual as well as state information of the environment which is extracted while navigating or interacting with it. Humans, exhibit this quality of multi-modal perception and cognition, which helps them decipher the surroundings in a way that they are even able to navigate unseen environments. Moreover, humans are able to perform such navigation and classification as a secondary task, while the goal of such movement can be varied. Example scenarios would include identification of different areas while navigation, using vision and tactile sensing to understand the current state of a surface or object. Another such example is performing status checks while navigating an unseen construction space. Identification-on-the-fly incorporates this ability of comprehending the unseen environment as an almost intuitive capability (performed as a secondary task) into autonomous robots, thereby taking them one step closer to human-like intelligence. 

In this paper, an identification-on-the-fly approach is utilized to address problems associated with coarse identification of human-built indoor spaces while navigating through them. This is accomplished based on an intuitive assumption that each class of space has its own unique signature. Moreover, common spaces exhibit similar patterns as they are built for specific purposes, such as staircases, corridors, etc. Hence, these unique signatures can be generalized throughout indoor spaces to learn and recognize the class of spaces for unseen environments too. 

\subsection{Indoor Construction Spaces}

Indoor construction spaces are unique environments in the sense that they have both static and dynamic elements. While the structure and walls may not change significantly over the course of time, the dynamic objects such as furniture, fittings, etc. can change drastically even over a short period of time. These changes pose a challenge to most autonomous system which rely on precise and real-time mappings. However, the coarse signature of the space remains rather constant. In this study, we leverage the overall signature of a space for coarse classification of the space.

\subsection{Clutter-Slices}
Clutter is the class of things which add context to a room. A room is primarily just four walls, however, if there are stairs in it then it becomes a staircase. Similarly, if the stairs are replaced by desks, it becomes a working space. Hence, there is inherent information, albeit coarse, in the distribution of objects in a four wall enclosure. Moreover, there is also information in the structure of the placement of the four walls. A corridor and an office, both have four walls but the structure is inherently dissimilar. The clutter-slices method leverages this inherent information in the distribution of objects and the basic structure of the enclosed spaces to classify the human-built environments. 

\begin{figure}[thpb]
    \centering
    \includegraphics[height=0.6\linewidth]{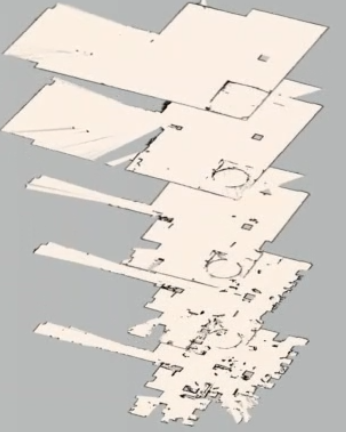}
    \caption{2D lidar scans of a room at multiple heights} 
 
    \label{fig:2dlidarmap}
\end{figure}

Clutter includes both static (wall fittings, doors, pillars, sinks) and dynamic objects (tables, chairs, table-top equipment, humans, cabinets). These objects occupy the scans with respect to their position in the environment. At different heights, different objects appear on the scan relative to their position in the environment, as illustrated in Fig. \ref{fig:2dlidarmap}. Based on the information available from the clutter-slices, different indoor facilities can exhibit unique distributions. 

Clutter slices do not just coarsely map the area, but they also coarsely localize the observer in the scene. The information from clutter slices enables abstraction of details such as the closeness of the observer to ceiling or ground and to the nearby walls. This information can be used to estimate the pose of the observer in the scene and subsequently map their trajectory. 

\section{Clutter Slices Dataset}
\begin{figure}[thpb]
    \centering
    \includegraphics[width=0.7\linewidth]{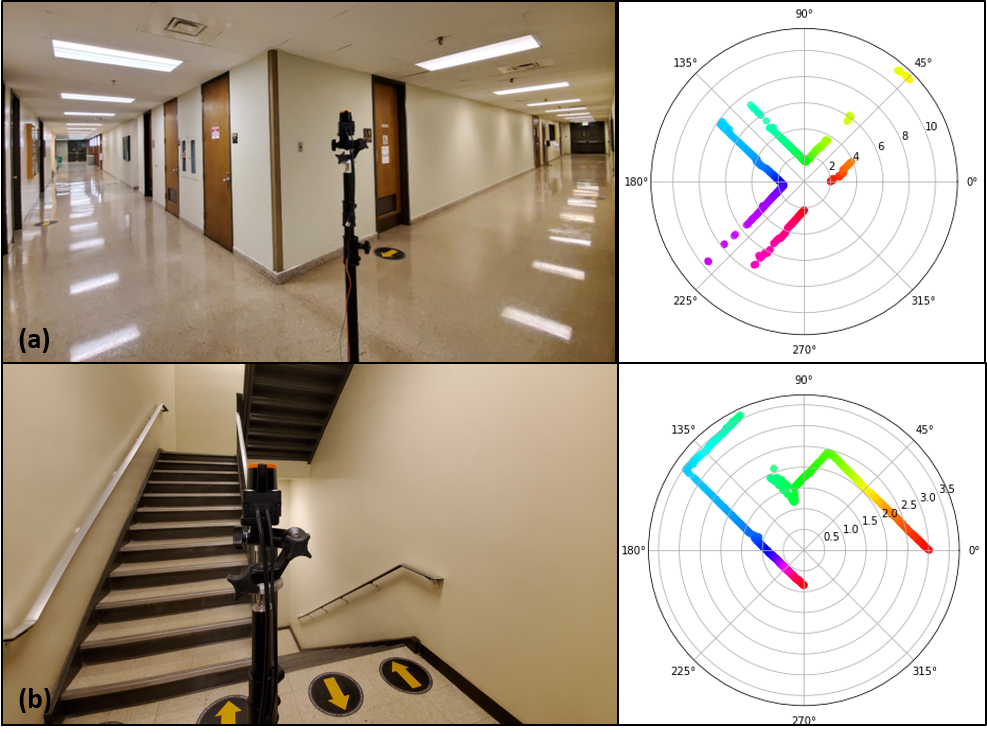}
    \caption{Images and respective 2D LiDAR plots of indoor spaces with the sensor capturing scans of (a)Corridor and (b) Staircase.} 
    \label{fig:cs2}
\end{figure}

Robust identification of construction spaces, especially indoor spaces, needs intelligent models that can comprehend the environment efficiently. The first step in building such models is creating adequate datasets for training. Hence, we present a diverse  dataset of real-life indoor spaces. The clutter slices dataset is a collection of scans of common indoor spaces, such as corridors, staircases, restrooms, and large shared spaces (including cafeterias, common areas, and shared working offices), as shown in Fig. \ref{fig:cs2}. This is a fully annotated dataset which enables models to learn the distribution of clutter in such common areas, and thereby contributes to efficient recognition of spaces. 

The Clutter Slices dataset was created by taking two-dimensional (2D) LiDAR scans of areas such as restrooms, staircases, shared spaces and corridors around the various buildings of Purdue University. We chose a LiDAR sensor for this data collection as it is one of the most widely used sensors in navigation and mapping in robotic vision. Using this sensor, we measure spatial distribution $270{^\circ}$ around a point, as shown in Fig. \ref{fig:cs2}. The maximum range of this sensor is 30 meters. Various positions around the space were used for the data collection to ensure a holistic capture of data. The height of the data collection was varied in steps of 1 meter.  

\begin{figure}[thpb]
    \centering
    \includegraphics[width=0.7\linewidth, height=0.4\linewidth]{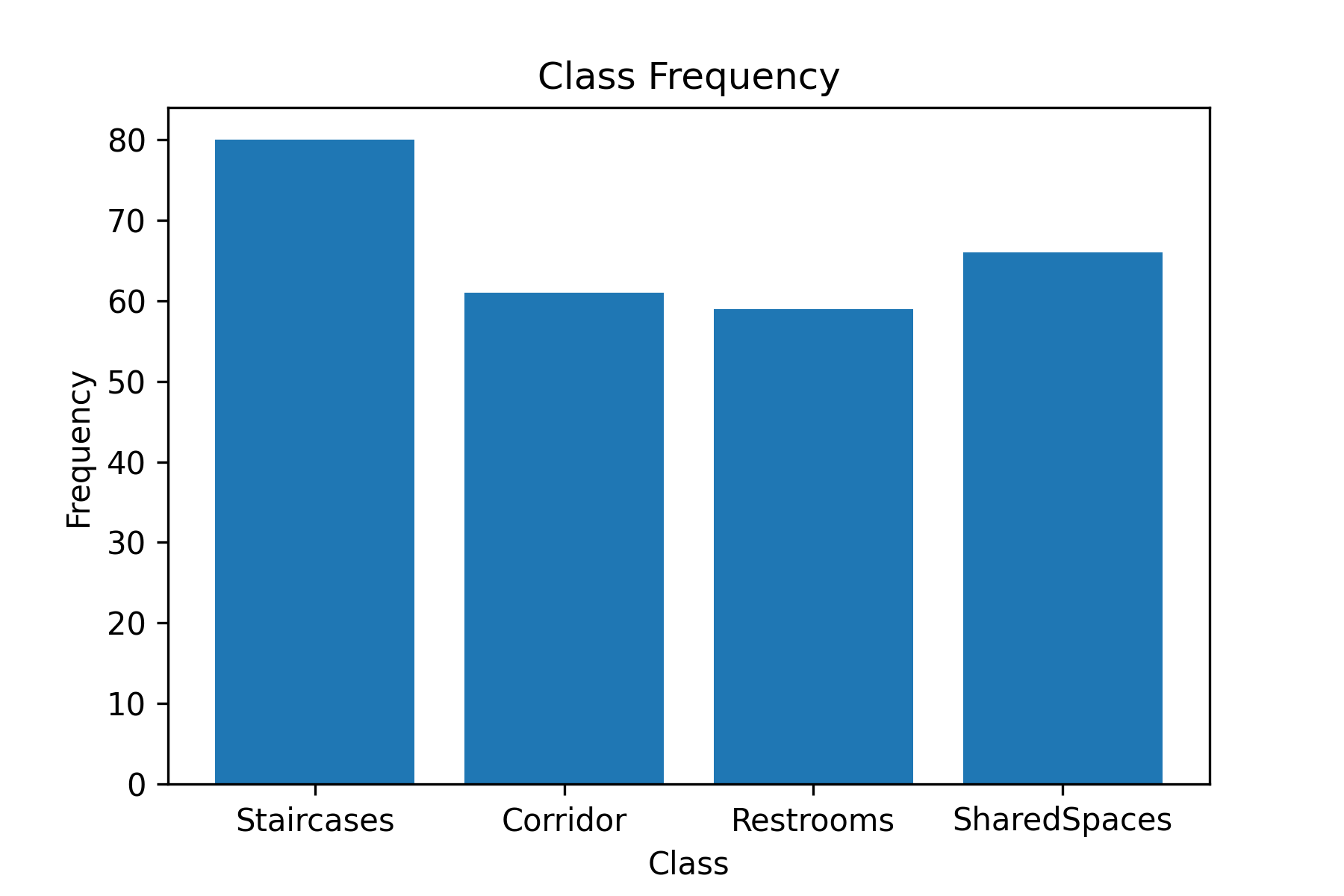}
    \caption{Frequency Distribution of Classes of Clutter Slices Dataset} 
    \label{fig:classdistribution}
\end{figure}

There are four classes in the Clutter Slices dataset: corridors, staircases, restrooms, and large shared spaces. These classes are common to most indoor construction areas and hence are useful for researchers in future work. The distribution of instances of these classes in the dataset are shown in Fig. \ref{fig:classdistribution}. The dataset is publicly available at https://github.com/CRLPurdue/Clutter\_Slices \cite{clutterdataset}. 

\section{Clutter Slices Pipeline}
The clutter slices approach with identification-on-the-fly aims to understand inherent patterns in the data, rather than relying on explicit feature engineering. Hence, by just using the distances of clutter around a point, we derive a clutter slice at a fixed height. A stack of these slices would build the clutter signature of the space. However, the goal here is to understand the strength of just a single clutter slice in deriving the class of a space. Therefore, we use a single 2D scan of the space to understand the distribution of clutter and subsequently, classify it. 

\begin{figure*}[!thpb]
    \centerline{
    \includegraphics[width=0.55\paperwidth, height=0.2\linewidth]{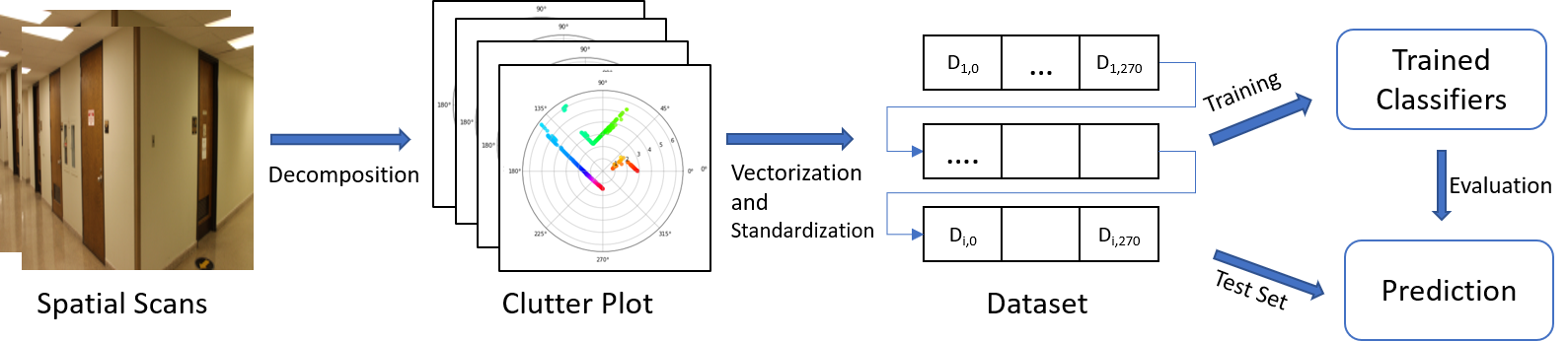}}
    \caption{Flowchart of Clutter Slices Pipeline}
    \label{fig:Flowchart}
\end{figure*}

In the clutter slices pipeline, the input 2D scan is translated to distances around the point. This allows for use of multiple sensors, as a variety of sensors such as LiDARs, cameras, and infrared sensors can be used to get the distance measurements. These distances are then vectorized  as the feature space $D{_i}$, wherein $D{_i} = [D{_i}{_,}{_0}, ..., D{_i}{_,}{_2}{_7}{_0}]$. The labels for this feature space are defined as $y{_i}$ where $i \in[0,3]$ for the clutter slices dataset. The feature space is then scaled using Box-Cox power transformations to standardize the data. The prepared data is then input to the classifiers. In this study, we used six classifiers which are widely used in machine learning: Random Forests, Logistic Regression, Support Vector Machines, AdaBoost, Artificial Neural Network, and Convolutional Neural Network. These classifiers present a baseline on the clutter slices dataset, and prove its effectiveness.

\section{Experiments and Results}
The validation of the proposed pipeline on the clutter slices dataset using the selected classifiers is presented in this section. We first present the experimental setup, including the hyperparameters selected for the classifiers, and consequently, present the performance in terms of accuracy, precision and recall for the classifiers. 

\subsection{Experimental Setup}

The experiments were conducted with the Clutter Slices dataset using the described pipeline with six classification models. Since this is a multi-class classification task, the dataset was stratified and shuffled, then split into a train and test set with an 80-20 ratio. We followed a five fold cross validation to ensure coverage of the entire dataset. The scikit-learn implementation of Random Forests (RF), Adaboost, Suppport Vector Machine (SVM), and Logistic Regression (LR) were all used \cite{scikit-learn}. A total of 100 estimators were used for RF with the total depth of 100. In case of Adaboost, the number of estimators used were 200. The polynomial kernel was used for SVM.

The architecture of the artificial neural network (ANN) constitutes of six fully connected dense layers. The number of units in the layers are: 481,364, 256, 125, 50 and 4. The last layer has Softmax activation with rectified linear units (ReLU) activation being used for the previous layers. We also incorporated two dropout layers in this network. The architecture of the convolutional neural network (CNN) comprises of two convolutional layers followed by a MaxPooling layer and three dense, fully-connected layers. The dense layers have 125, 50 and 4 units, respectively. Dropout and input flattening layers were also used in this network. Softmax activation was used at the last layer with ReLU being used in all others.  The CNN and the ANN, both used the Adam optimizer with a learning rate of 0.01. The categorical cross-entropy was used as a measure of loss. Both neural network models were trained for 30 epochs with a mini-batch size of 32. 

The training and testing was conducted on a computer with 32GB RAM, NVIDIA GTX 1080 Ti GPU and Intel Core i9 CPU. 

\begin{table}[!h]
\caption{Accuracy on test set for the Clutter Slices Dataset}
\begin{tabular}{@{}ccccccc@{}}
\toprule
\multirow{2}{*}{Classifiers} & \multicolumn{5}{|c|}{Cross validation Accuracy} & \multirow{2}{*}{Overall Accuracy} \\ \cmidrule(lr){2-6}
 & \multicolumn{1}{|c|}{1st Fold} & \multicolumn{1}{c|}{2nd Fold} & \multicolumn{1}{c|}{3rd Fold} & \multicolumn{1}{c|}{4th Fold} & \multicolumn{1}{l|}{5th Fold} &  \\ \midrule
\multicolumn{1}{|c|}{RF} & \multicolumn{1}{c|}{$0.907$} & \multicolumn{1}{c|}{$0.88$} & \multicolumn{1}{c|}{$0.94$} & \multicolumn{1}{c|}{$0.96$} & \multicolumn{1}{c|}{$0.94$} & \multicolumn{1}{c|}{$0.928 \pm 0.03$} \\ \midrule
\multicolumn{1}{|c|}{AdaBoost} & \multicolumn{1}{c|}{$0.57$} & \multicolumn{1}{c|}{$0.396$} & \multicolumn{1}{c|}{$0.53$} & \multicolumn{1}{c|}{$0.60$} & \multicolumn{1}{c|}{$0.37$} & \multicolumn{1}{c|}{$0.495 \pm 0.09$} \\ \midrule
\multicolumn{1}{|c|}{SVM} & \multicolumn{1}{c|}{$0.83$} & \multicolumn{1}{c|}{$0.88$} & \multicolumn{1}{c|}{$0.867$} & \multicolumn{1}{c|}{$0.924$} & \multicolumn{1}{c|}{$0.886$} & \multicolumn{1}{c|}{$0.88 \pm 0.03$} \\ \midrule
\multicolumn{1}{|c|}{Logistic Regression} & \multicolumn{1}{c|}{0.759} & \multicolumn{1}{c|}{0.849} & \multicolumn{1}{c|}{0.83} & \multicolumn{1}{c|}{0.79} & \multicolumn{1}{c|}{0.849} & \multicolumn{1}{c|}{$0.82 \pm 0.035$} \\ \midrule
\multicolumn{1}{|c|}{CNN} & \multicolumn{1}{c|}{0.907} & \multicolumn{1}{c|}{0.905} & \multicolumn{1}{c|}{0.94} & \multicolumn{1}{c|}{0.96} & \multicolumn{1}{c|}{0.96} & \multicolumn{1}{c|}{$0.936 \pm 0.03$} \\ \midrule
\multicolumn{1}{|c|}{ANN} & \multicolumn{1}{c|}{0.87} & \multicolumn{1}{c|}{0.87} & \multicolumn{1}{c|}{0.925} & \multicolumn{1}{c|}{0.96} & \multicolumn{1}{c|}{0.89} & \multicolumn{1}{c|}{$0.90 \pm 0.04$} \\ \midrule
 &  &  &  &  &  &  \\ \bottomrule
\label{table:acc}
\end{tabular}
\end{table}

\subsection{Results}
The tests were performed using the Clutter Slices dataset. The accuracy of the six classifiers for each fold, along with the overall accuracy is presented in Table \ref{table:acc}. The results indicate that the clutter slices dataset is able to present enough information for recognition of classes, even with just a single scan as input. While random forests, CNN, and ANN models showed more than 90\% accuracy, models like SVM and Logistic regression also showed good performance with very little hyper-parameter tuning. The low accuracy of Adaboost can be attributed to over-fitting by the model. 

\begin{figure}[!ht]
    \centerline{
    \includegraphics[width=0.55\paperwidth, height=0.77\linewidth]{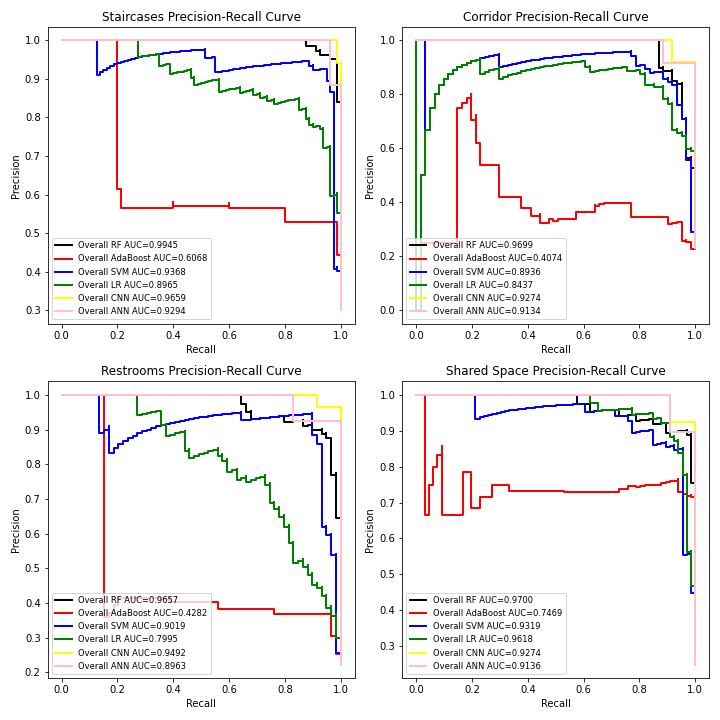}}
    \caption{Overall classifier performance} 
    \label{fig:testingresults}
\end{figure}

Figure \ref{fig:testingresults} shows the class-wise precision recall curves for the overall performance of the six classifiers. These curves highlight that the models were able to identify classes Staircases and Shared Spaces without much loss, but Restrooms and Corridors were showing overlap with other classes. The overlap can be intuitively explained as restrooms can have characteristics similar to shared spaces. Nevertheless, despite these challenges, the area-under-the-curve (auc) values prove the performance of these models. 

\newpage
\section{Conclusion}

In this paper we introduce the identification-on-the-fly approach to imbue \\ human-like intelligence into robotic systems. The proposed clutter slices approach leverages the unique signatures of common indoor spaces for coarse classification. The initial validation of the clutter slices approach is performed on the dataset using 2D LiDAR sensor. Further, we present a scalable pipeline that supports this approach. The pipeline is flexible enough to accommodate varied classifiers. We used some of the widely used classifiers such as random forests, logistic regression, and neural network models to establish a baseline for the dataset. A maximum accuracy of 93.6\% was achieved with this approach without significant hyperparameter tuning. The precision-recall plots show the convergence of the models in recognizing the classes of spaces. 

The clutter slices approach captures the unique signatures of common indoor spaces and proves the potential of this approach in their coarse classification. Nevertheless, the clutter slices approach is not sensor specific and can be potentially generalized across domains. In the future, this approach of identification-on-the-fly can be an essential tool for perceiving and assessing surroundings of intelligent autonomous agents. Clutter slices is one implementation of the identification-on-the-fly method used for coarse classification of indoor spaces, adding contextual information to the robot perception. However, there are endless opportunities to perform identification-on-the-fly to understand the surrounding while still identifying potential dangers and outcome of future actions. 

\section*{Acknowledgment}
This work was supported, in part, by the Dept. of Energy, the NSF Center for Robots and Sensor for the Human Well-Being (RoSe-HUB) and by the National Science Foundation under grant CNS-1439717 and the USDA under grant 2018-67007-28439.
The authors greatly acknowledge the contribution of coffee for the actualization of this work.

\bibliographystyle{splncs04}
\bibliography{mybib}

\end{document}